# A Novel Machine Learning Classifier Based on Genetic Algorithms and Data Importance Reformatting

A. K. Alkhayyat[a] and N. M. Hewahi[b]

[a]Big Data Science and Analytics Program, College of Science, University of Bahrain, Bahrain (e-mail: aisha.alkhayat@gmail.com); [b]Department of Computer Science, College of Information Technology, University of Bahrain, Bahrain

## Abstract

In this paper, a novel classification algorithm that is based on Data Importance (DI) reformatting and Genetic Algorithms (GA) named GADIC is proposed to overcome the issues related to the nature of data which may hinder the performance of the Machine Learning (ML) classifiers. GADIC comprises three phases which are data reformatting phase which depends on DI concept, training phase where GA is applied on the reformatted training dataset, and testing phase where the instances of the reformatted testing dataset are being averaged based on similar instances in the training dataset. GADIC is an approach that utilizes the exiting ML classifiers with involvement of data reformatting, using GA to tune the inputs, and averaging the similar instances to the unknown instance. The averaging of the instances becomes the unknown instance to be classified in the stage of testing. GADIC has been tested on five existing ML classifiers which are Support Vector Machine (SVM), K-Nearest Neighbour (KNN), Logistic Regression (LR), Decision Tree (DT), and Naïve Bayes (NB). All were evaluated using seven open-source UCI ML repository and Kaggle datasets which are Cleveland heart disease, Indian liver patient, Pima Indian diabetes, employee future prediction, telecom churn prediction, bank customer churn, and tech students. In terms of accuracy, the results showed that, with the exception of approximately 1% decrease in the accuracy of NB classifier in Cleveland heart disease dataset, GADIC significantly enhanced the performance of most ML classifiers using various datasets. In addition, KNN with GADIC showed the greatest performance gain when compared with other ML classifiers with GADIC followed by SVM while LR had the lowest improvement. The lowest average improvement that GADIC could achieve is 5.96%, whereas the maximum average improvement reached 16.79%.

**Keywords:** machine learning; genetic algorithms; classification; data importance

## 1. Introduction

ML has enabled industries to be more efficient and cost-effective. With the exponential growth of data, a need has arisen for ML algorithms with better prediction capabilities. This may be accomplished by applying optimization methods (OM), which can improve the efficiency of the used ML algorithms. This can help diverse fields to accurately predict the occurrence of certain conditions that are of paramount value to them.

Because to the nature of the data, some classifiers often perform poorly and achieve low performance. Data scientists may improve the performance of ML algorithms by altering the hyperparameters or carrying out additional pre-processing steps. However, there are times when the algorithm's performance remains low, suggesting that the problem lies within the data itself.

In this paper, a novel classification algorithm named (GADIC) is proposed that is based on GA and DI reformatting for the ultimate goal of enhancing the performance of the classifiers. GADIC uses existing classifiers to measure its performance. The main approach relies on altering the data representation in accordance with the DI technique presented in Hewahi (2019) and then applying GA as an OM to tune and adjust the data. The data reformatting is performed based on the influence of certain values within the attributes in affecting the produced output. Applying GA in both training and testing phases on the reformatted data to tune it and avoid any loss is a key step to enhance the performance of the classifiers. GADIC is anticipated to be capable of overcoming some of data related impediments that might prevent the classifier from performing well, hence, elevating the performance of the classifier.

## 2. Background

The three core elements of GADIC are ML classifiers, GA, and data reformatting.

### 2.1. Classifiers

Under this section we briefly present the ML algorithms used in our GADIC proposed approach.

**SVM**

SVM is a supervised method that its decision function is an optimal hyperplane that separates classes of the observations based on their features. Accordingly, the best hyperplane is the one that maximizes the margin between classes by utilizing the support vector points. Recently, the community of ML draw more attention to SVM due to its exceptional generalization capability and discriminative power as it is able to handle high-dimensional data (Cervantes et al., 2020; Pisner & Schnyer, 2020). The advantage of SVM is that it can handle both semi-structured and structured data by using a kernel function to transform the input space into a higher dimensional space. However, SVM can be computationally expensive especially with large datasets due to the extended training time. Furthermore, it has the difficulty of selecting the right kernel function (Ray, 2019).

**KNN**

KNN is a non-parametric algorithm which essentially reflects the number (k) of neighbours in the training data that are closest to a specific testing point. The most crucial step in KNN is choosing the proper k value. The nearest neighbours are selected based on Euclidean Distance Function. Eventually, the label of the given testing point is determined by a majority vote of the chosen k nearest neighbours Despite being a simple and flexible algorithm, KNN has a high computation cost and is sensitive to the selection of k value as it is the only parameter (Ray, 2019; Taunk et al., 2019; Cheng et al., 2014).

**LR**

LR is a statistical method for binary and multiclass classification problems in ML as a sort of regression (Connelly, 2020; Maalouf, 2011). It is used to discover the relationship between a dependent variable and one or more independent variables in addition to predict a binary outcome based on those independent variables. The fundamental notion of LR is that it provides probability estimations that range from 0 to 1 (Maalouf, 2011). The main drawback of this algorithm is its assumption of linearity between input and output. Yet, it is easy to implement, train, and interpret (Ray, 2019; Maalouf, 2011).

**DT**

DT is a tree-based classification algorithm in which data is recursively splitting based on a specific threshold. Starting with a root node that represents the complete dataset, DT splits the data into smaller subsets based on the features and their values until final leaves are reached. In the final leaves, each subset consists only of instances of a single class (Jijo & Abdulazeez, 2021; Ray, 2019; Priyam et al., 2013). DT is renowned for its ease in implementation and handling quantitative, qualitative, and missing data. Nonetheless, DT has flaws such as instability and difficulty in controlling the tree size (Jijo & Abdulazeez, 2021).

**NB**

NB is a probabilistic method that implements Bayes Theorem to estimate the probabilities of the dataset's values. This is done by calculating the frequency of their occurrence. The assumption that the attributes are independent in NB is problematic. However, it is known for its scalability, ability of handling missing values and multi-class problems, as well as its suitability to handle high dimensional data (Peling et al., 2017; Taheri & Mammadov, 2013). Despite that, NB's simplicity may result in lower performance compared to other models that have been adequately trained and tuned (Ray, 2019).

### 2.2. GA

GA lies in the family of evolutionary algorithms, a type of meta-heuristic optimization algorithm, which are computational search methods that draws their inspiration from natural selection (Vié et al., 2021; Savio & Chakraborty, 2019; Dahiya & Sangwan, 2018). The Darwinian theory of "Survival of the fittest" is the cornerstone of the GA (Katoch et al., 2021; Savio & Chakraborty, 2019). GA operations rely on randomness to set the values of their parameters (Alam et al., 2020).

GA is carried out through a series of standardized processes. In this method, the problem space is abstracted as a population of individuals, and it attempts to discover which individual is the most fit by constructing generations repeatedly. The fitness function, which identifies the criterion for rating prospective populations and choosing them for inclusion in the following generation, determines the optimal solution to a given problem (Haldurai et al., 2016; Haldar et al., 2014).

Selection, crossover, and mutation are the three main genetic operators that are consecutively applied with specific probabilities to every individual during each generation until a predefined termination criterion is met (Haldurai et al., 2016). Based on the fitness values of the individuals in the population, the selection operator determines which individuals of the population will be selected for reproduction in the following generation. The crossover operator involves exchanging genes between two selected individuals (by selection operator) to generate new offspring with genes from both parents, while mutation operator randomly alters some of the genes in an individual.

GA has demonstrated its effectiveness as a robust OM. This, together with its flexibility, simplicity in comprehending, capability of finding the global optimum solution, and ability of solving complex optimization problems that are difficult to solve using other OM, contributes to its increasing recognition and adoption (Haldurai et al., 2016; Tabassum & Mathew, 2014). With concentrating on bio-inspired operators, GA is widely employed to provide high quality solutions to optimization and search problems. GA can settle problems in the real world by replicating this process (Alam et al., 2020; Dahiya & Sangwan, 2018).

GA, though, encounter significant challenges that have constrained its uses. One thing is its high computational cost. For satisfactory results, GA may need to run for a lengthy time. Additionally, a larger population or complex problems typically cause a considerable slowdown of GA. Another thing is the difficulty in determining the suitable parameter's settings i.e. selecting the initial population, fitness function, and degree of crossover and mutation (Katoch et al., 2021; Vié et al., 2021).

### 2.3. Data Reformatting

The data reformatting method employed in this paper was adopted from research conducted by Hewahi (2019). According to Hewahi (2019), the method is formulated on how frequently a specific attribute value is replicated across all dataset instances. Based on potential changes in the attribute's values across the dataset, the importance of each input for each attribute is determined.

The essence of the method is that an attribute's relevance in influencing the output generated decreases in proportion to how small its values changes. Likewise, the more an attribute's values change across the dataset, the more important it is to the process of producing a certain outcome. The importance of each input is calculated using equations established by Hewahi (2019). All data inputs in the dataset are eventually replaced with their estimated importance values.

## 3. Related Work

### 3.1. Optimization Methods

The optimization of ML algorithms can be carried out in a variety of methods, including Feature selection (FS), Parameter Optimization (PO), regularization, model selection, and ensemble methods (Brownlee, 2021; Bains, 2020; Brownlee, 2020). A summary of the reviewed GA-based ML algorithm optimizations is presented in Table 1.

All studies reported in Table 1 revealed an improvement in classification performance when applying GA as an OM with ML algorithms. Most studies showed a high rate of accuracy growth. The accuracy reached 100% when KNN was combined with GA using the lung cancer dataset (Maleki et al., 2021). Among various algorithms that were integrated with GA, it was observed that NN, XGBoost, and RF algorithms achieved the highest increase in performance. Furthermore, the performance improvement of SVM and DT is greater than that of NB and LR (Demir & Şahin, 2022; Kanwal et al., 2021; Rahmadani et al., 2018). In conjunction with GA, applying ensemble methods instead of a single algorithm led to yet more improvement in performance (Ashri et al., 2021).

Table 1. Summary of research applied ML algorithms in conjunction with GA

| Reference | Applied ML | OM | Dataset |
|---|---|---|---|
| Azad et al. (2022) | DT | FS | UCI Pima Indian diabetes |
| Demir and Şahin (2022) | SVM, RF, XGBoost | FS + PO | Vs dataset |
| Ashri et al. (2021) | Ensemble DT and RF | FS | UCI heart disease |
| Kanwal et al. (2021) | SVM, Deep Learning, NB, NN, LR | FS | UCI heart disease datasets (Cleveland and Statlog) |
| Maleki et al. (2021) | KNN | FS | Lung cancer dataset from data world site |
| Soumaya et al. (2021) | SVM | FS | Parkinson |
| Farid et al. (2020) | C4.5 | FS | UCI Wisconsin |
| Safitri and Muslim (2020) | NB | FS | UCI German credit dataset |
| Rahmadani et al. (2018) | DT, NB | FS | UCI datasets (iris, glass, credit) |
| Pawlovsky and Matsuhashi (2017) | KNN | FS | UCI breast cancer |
| Aly (2016) | LR | FS + PO | UCI Pima Indian diabetes |
| Choubey et al. (2016) | NB | FS | UCI Pima Indian diabetes |

### 3.2. Data Reformatting Techniques

A new encoding approach based on existing encoding techniques was proposed by Tabassum et al. (2020) to handle categorical variables in a dataset. The developed approach, known as "one-hot-frequency", combines one-hot and frequency encoding methods. A number of ML algorithms' classification performance was improved by the proposed approach. The best performance, though, came from RF followed by DT and ANN.

A strategy for image quality enhancement was introduced by Hung et al. (2021). This method can enhance image features by incorporating technologies such as noise reduction, sharpening, and color channel adjustment. This method was combined with Deep Learning (DL) model to improve the classification performance.

Hewahi (2019) proposed a new method for data reformatting in which all the dataset inputs are eventually replaced based on their importance in affecting the produced output. The breast cancer, iris, and diabetes datasets from the UCI ML repository were subjected to the proposed data reformatting technique. Employing this technique in conjunction with NN pruning algorithm has yielded classification performance that is superior to that of other NN pruning algorithms in most of the tested datasets.

## 4. The Proposed Approach: GADIC Algorithm

GADIC comprises three phases which are data reformatting phase which depends on DI concept, training phase where GA is applied on the reformatted training dataset, and testing phase where the instances of the reformatted testing dataset are being averaged based on similar instances in the training dataset. GADIC is tested on five ML classifiers namely, SVM, KNN, LR, DT, and NB. Accuracy and f1-score are utilized to evaluate the effectiveness of GADIC and compare the results. The steps of GADIC are illustrated in Figure 1. The GADIC algorithm is given below.

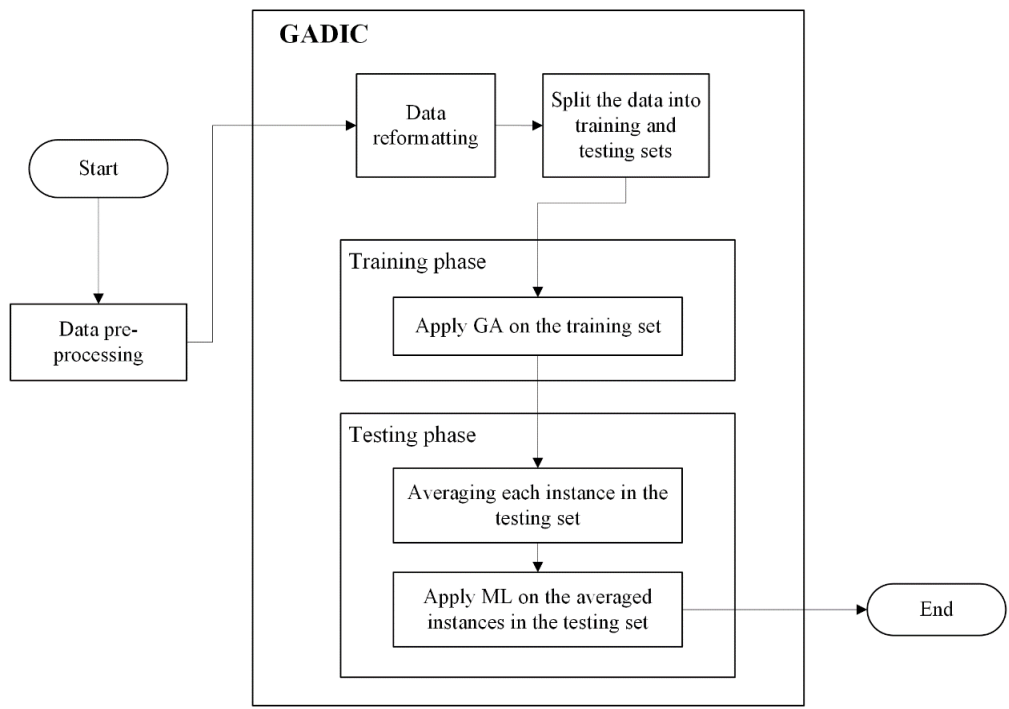

Figure 1. Flowchart of the proposed algorithm

| Algorithm: GADIC Algorithm |
|---|
| **Start** |
| **# Data reformatting phase** |
| **Input of phase:** random sample size s, number of repetitions w, number of intervals $pt_i$ |
| **For** (i in range (attributes in the dataset)) |
|     Select s |
|     **If** (datatype = integer) |
|         Calculate proportion of possible values $r_{im}$ using Eq. 1 |
|         Calculate $\beta i$ using Eq. 2 |
|     **Elseif** (datatype = float) |
|         Calculate size of interval $int_i$ using Eq. 5 |
|         Calculate proportion $r_{im}$ using Eq. 6 |
|         Calculate $\beta i$ using Eq. 2 |
|     Repeat w times to find average$\beta i$ using Eq. 3 |
|     Compute importance $I_{iy}$ using Eq. 4 |

    Replace attribute values with their $I_{iy}$

**Output of phase:** reformatted dataset based on data reformatting phase

# Data splitting

Split the reformatted dataset into training and testing datasets

# Training phase

**Input of phase:** termination generation, crossover probability cp, crossover point, mutation probability mp, mutation rate, mutation value, training dataset

Set count = 0

Set initial population = training dataset

**While** (count < termination generation)

    Select 2 instances as parents from population randomly

    Apply crossover on parents based on cp

    Apply mutation on offspring obtained from crossover based on mp

    Compute fitness function for the new offspring

    **If** (fitness = True)

        Replace parents with offspring in the initial population

    **Else**

        Don't replace parents with offspring in the initial population

    Count+=1

**End while**

**Output of phase:** final training dataset

# Testing phase

**Input of phase:** number of closest rows n, the final training dataset

**For** (i in range (instances in the testing dataset))

    Find n closet instances in the training dataset using Euclidean distance function

    **For** (i in range (attributes in the testing dataset))

        Calculate average attribute value of the retrieved closest instances using Eq. 8

Apply ML classifier on the averaged instances

Assess the performance

**Output of phase:** performance measures results

**End**

### 4.1. Data Reformatting Phase

In this phase, based on Hewahi (2019), the importance of each attribute input in producing certain output in a dataset will be calculated. Changing factor β for each attribute will be used to calculate the importance of each attribute input. Each input in that attribute will be replaced by its importance value. All steps will be carried out for each attribute independently in the dataset in order to compute the importance of its inputs and replace their values in accordance with their importance. The main steps of the data reformatting phase are as follow:

1. Select a random sample of the given dataset where *s* is the number of randomly selected instances.
2. In the case where the attribute has categorical or integer datatypes, the following steps will be followed:

    a. Each attribute *i* has $p_i$ possible values. For each possible value in the attribute *i*, the proportion of every attribute input $a_i$ is calculated as shown in Equation (1):

    $$r_{im} = \frac{v_{im}}{s}, \quad (1)$$

    where,

    $v_{im}$ is the number of m-equivalent values in the random sample for $a_i$.

    b. All $r_{im}$ values calculated by Equation (1) is then appended to a set $R_i$ that will contain all proportion values of $a_i$.

    $$R_i = \{r_{im1}, r_{im2}, r_{im3}, ...., r_{imp}\}$$

    c. Then, the set $R_i$ is utilized to calculate the changing factor $\beta_i$ for each attribute *i* using Equation (2):

    $$\beta_i = \frac{maxR_i - minR_i}{2}, \quad (2)$$

    where,

    $maxR_i$ is the maximum value in $R_i$,

    $minR_i$ is the minimum value in $R_i$.

    d. To reduce the risk of bias in the random selection process, the step of the random selection of instances and the subsequent stages are

repeated for $w$ times. Then, the average of $\beta_i$ is calculated using Equation (3):

$$avg\beta_i = \frac{\beta_{i1} + \beta_{i2} + \cdots + \beta_{iw}}{w}, \tag{3}$$

where,

$w$ is the number of repetitions of the random selection step.

e. Calculate the importance of each input value $a_i$ for each instance as given in Equation (4):

$$I_{iy} = X_{iy}(1 - \beta_i), \tag{4}$$

where,

$X_{iy}$ is the input value for the attribute $i$ of instance $y$,

$I_{iy}$ is the importance of $X_{iy}$.

f. Replace each input value $a_i$ in attribute $i$ with its calculated importance value $I_{iy}$ in the dataset.

3. While in the case where the attribute has real numbers (float datatype), only parts a and b will be different than those of integer or categorical data (step 2) due to the nature of data inputs that requires different set of calculations. Remaining steps for real data (c until f) will be the same as identified earlier in step 2. Steps a and b for real data are as follow:

a. Calculate the size of interval $int_i$ for the attribute inputs $a_i$ of the attribute $i$ as given in Equation (5):

$$int_i = \frac{maxValue_i - minValue_i}{pt_i}, \tag{5}$$

where,

$maxValue_i$ is the maximum value in attribute $i$,

$minValue_i$ is the minimum value in attribute $i$,

$pt_i$ is an integer number that specifies the number of data partitions for the dataset in which the more is this number, the less is the size of the interval and vice versa.

b. Find the first and last value of every interval based on the calculated $int_i$. After that, assign each attribute input $a_i$ in the attribute $i$ to the interval to which it belongs. Then calculate the proportion of each $a_i$ using Equation (6):

$$r_{im} = \frac{c_{im}}{s}, \tag{6}$$

where,

$c_{im}$ is number of values in a certain interval within the attribute $i$ in the random sample $s$.

The calculated changing factor $\beta_i$ is a value between 0 and 0.5. Smaller values of $\beta_i$ indicates that the value of $a_i$ is fluctuating and thus have a greater impact on producing the output. Whereas higher $B_i$ values indicate that the value of $a_i$ does not fluctuate significantly and may thus have less impact on producing the output.

### 4.2. GA

In this phase, the training dataset will be subjected to GA operators in order to fine-tune the data. The whole training dataset constitutes the population in GADIC. Also, the fitness function is built to find the attribute values that will yield to the best performance. The steps in this phase will be repeated until the termination condition specified at the beginning is met. The sequence of the steps of GA phase in GADIC is illustrated in Figure 2.

The steps of GA phase in the proposed algorithm are as follow:

1. Set a termination condition for GA to stop and return the final results. In the proposed algorithm, the termination condition is satisfied when a number of generations $g$ is reached. This number is specified before applying any GA operators.
2. Select two instances randomly from the training set as parents to perform the GA operators on them.
3. Apply single point crossover on the selected parents. This type of crossover is carried out based on two main parameters:
   a. Crossover probability which is a parameter that regulates the possibility that crossover will happen between the parents. It is more

probable that crossover between the parents will take place if the crossover probability is set to a high number and vice versa.

b. Crossover point which is a location selected randomly in the instance. The two parents will be cut at that point and the data to the right of this location are swapped between the two parents to produce two offspring.

In this proposed algorithm, the crossover will be applied only on the independent attributes. The class labels will remain untouched. The crossover is performed at the selected crossover point only if it satisfies the crossover probability criteria.

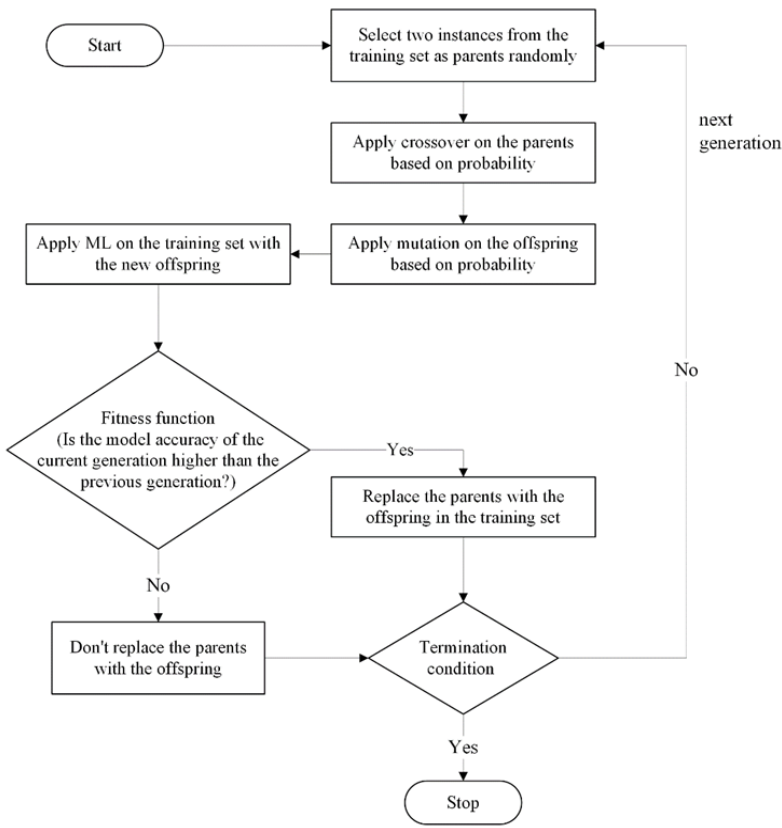

Figure 2. Flowchart of GA phase in GADIC

4. Apply mutation on the offspring produced in the previous step. The mutation is carried out based on three main parameters:

    a. Mutation probability which is a parameter that regulates the possibility that an instance will be subjected to the mutation process. The probability should ideally be set to a small number so that only a very small portion of the population mutates in each generation.

    b. Genes to be mutated in the offspring. A mutation rate that is specified at the beginning determines how many genes will be mutated while the locations of these genes are selected randomly.

    c. Mutation value which is a value chosen at random between 0 and 0.1. This value will be added to the values of genes that were selected in step b.

It is worth mentioning that the mutation is performed only if it satisfies the mutation probability criteria.

5. Apply the fitness function to the produced training set after crossover and mutation operators. The fitness function in the proposed algorithm is designed to assess whether applying GA operators on the reformatted dataset will yield to an increase in the performance or not. Only if the fitness function returns a True value, the offspring will replace their parents in the initial dataset. Otherwise, they will be discarded. The fitness function process is as follows:

    a. At the first generation, the fitness function will return True if the model accuracy after the crossover and mutation operators is higher than the initial model accuracy. If not, it will return False.

    b. At the subsequent generations, the fitness function will return True if the model accuracy after the crossover and mutation operators in that generation is better than the previous generation's model accuracy. Otherwise, it will return False.

### 4.3. Averaging Each Instance in the Testing Set

In this phase, the attributes values of each instance in the testing test is averaged using the attributes values of their similar instances in the training dataset. The following are the steps of the testing phase. These steps will be repeated for each instance in the testing dataset.

1. For instance *n* in the testing set, find the *h* instances in the training set that are closest to *n*, where *h* is the chosen number of the similar instances from the training set. The value of *h* will be selected based on trial and error to find the best value for it. Euclidean Distance function will be used to find the *h* similar instances from the training set to instance *n*. Such distance is calculated using Equation (7):

$$d_{i,j} = \sqrt{\sum_{p=1}^{k}(x_{ip} - x_{jp})^2},  \quad (7)$$

where,

$p=1, \ldots, k$ are the attributes,
$x_{ip}$ is the input value of attribute *p* in instance *i*,
$x_{jp}$ is the input value of attribute *p* in instance *j*.

2. Calculate the average value for each attribute in the similar instances independently. Equation (8) is used to find the average $avgAtt_i$:

$$avgAtt_i = \frac{v_1+v_2+\cdots+v_h}{h},  \quad (8)$$

where,

*v* is the input value of the specified attribute in the selected instances.

3. The average values $avgAtt_i$ obtained in the prior step for each attribute will be used in place of the values of instance *n* in the testing set.

## 5. Experiments

The datasets used in this paper were obtained from Kaggle and UCI ML repository. GADIC experiments were run on seven datasets which are Cleveland heart disease, Indian liver patient, Pima Indian diabetes, employee future prediction, telecom churn prediction, bank customer churn, and tech students.

The chosen datasets for the experiments ranged in size from a few hundred to 20,000 instances. All of them are classification datasets with either binary or multiclass class labels. Input values of various datatypes are present in these datasets. With the exception of the tech students dataset, which includes six class labels, all datasets have binary class labels. The only datasets with balanced or almost equal classes are Cleveland heart disease and tech students. Imbalances exist in the other five datasets.

The pre-processing steps employed in this paper include the elimination of irrelevant attributes, the removal of missing and duplicate data, and the application of encoding techniques on the categorical attributes.

Extensive experiments were conducted throughout every single phase to find the best parameters to be employed in each phase of GADIC. Python libraries were used in Jupyter notebook to conduct all experiments and compare the results.

In the data reformatting phase, the optimal parameter values for the datasets of sizes up to 5,000 records were 50 for the random sample size, 10 for the number of repetitions, and 4 for the number of intervals. While, for larger datasets, the parameter values were raised to 100 for the random sample size, 50 for the number of repetitions, and 10 for the number of intervals. As for data splitting of the datasets, 70/30 split produced the optimum performance across most datasets, hence it was used for all experiments.

In the training phase, the termination condition was set to the completion of 25,000 generations. Table 2 displays the parameter values for crossover operator that led to the best performance along with the number of generations it took to achieve that result. As for mutation operator, based on the recommendation by Datta (2023) and Savio and Chakraborty (2019) that the mutation probability should be set to a low number, all mutation parameters were set at low values. The mutation rate was set to maximum 0.2 randomly, the mutation probability was set to 0.3, and the mutation value was set to a random value between 0 and 0.1 for all experiments.

In the testing phase, the number of similar instances (n) to be used in the Euclidean distance function is the only parameter in this phase. For all datasets and classifiers, the n values that produced the highest performance in the experiments fell between 3 and 7.

Table 2. Parameters resulted with the best performance in GADIC

| Dataset | Classifier | Crossover point | Crossover probability | No. of generations |
|---|---|---|---|---|
| Cleveland Heart Disease | SVM | 3 | 0.6 | 24,768 |
|  | KNN | 3 | 0.6 | 13,208 |
|  | LR | 7 | 0.7 | 8 |
|  | DT | 4 | 0.7 | 6,815 |
|  | NB | 4 | 0.7 | 19 |
| Indian Liver Patient | SVM | 3 | 0.7 | 2,951 |
|  | KNN | 4 | 0.6 | 23,442 |
|  | LR | 4 | 0.6 | 10,531 |
|  | DT | 3 | 0.6 | 24,647 |
|  | NB | 3 | 0.5 | 1,558 |
| Pima Indian Diabetes | SVM | 3 | 0.7 | 24,595 |
|  | KNN | 4 | 0.6 | 5,530 |
|  | LR | 3 | 0.6 | 15,570 |
|  | DT | 3 | 0.7 | 6,942 |
|  | NB | 3 | 0.6 | 15,718 |
| Employee Future Prediction | SVM | 3 | 0.5 | 10,869 |
|  | KNN | 6 | 0.6 | 667 |
|  | LR | 7 | 0.7 | 5,735 |
|  | DT | 7 | 0.7 | 42 |
|  | NB | 7 | 0.6 | 5,608 |
| Telecom Churn Prediction | SVM | 7 | 0.5 | 521 |
|  | KNN | 4 | 0.6 | 43 |
|  | LR | 11 | 0.6 | 1,102 |
|  | DT | 9 | 0.6 | 179 |
|  | NB | 7 | 0.7 | 118 |
| Bank Customer Churn | SVM | 3 | 0.7 | 2,482 |
|  | KNN | 7 | 0.7 | 43 |
| | LR | 4 | 0.6 | 1,481 |
| | DT | 3 | 0.7 | 371 |
| | NB | 3 | 0.6 | 1,694 |
| Tech Students | SVM | 6 | 0.6 | 30 |
| | KNN | 5 | 0.7 | 7 |
| | LR | 7 | 0.6 | 59 |
| | DT | 4 | 0.7 | 5 |
| | NB | 6 | 0.7 | 14 |

## 6. Results

To determine the effectiveness of GADIC, the datasets were first evaluated using the conventional ML classifiers, and then GADIC was tested on the same datasets to compare the results. The performance was evaluated using accuracy and f1-score measures. Table 3 and Table 4 show the difference in accuracy before and after applying GADIC on the datasets.

Table 3. Accuracy before applying GADIC

| | SVM | KNN | LR | DT | NB |
|---|---|---|---|---|---|
| Cleveland Heart Disease | 67.77% | 67.77% | 85.55% | 77.77% | 88.88% |
| Indian Liver Patient | 71.17% | 68.23% | 74.70% | 61.17% | 56.47% |
| Pima Indian Diabetes | 75.75% | 74.45% | 77.05% | 68.83% | 77.48% |
| Employee Future Prediction | 65.61% | 78.15% | 70.84% | 81.73% | 68.98% |
| Telecom Churn Prediction | 73.41% | 76.91% | 79.85% | 72.93% | 75.87% |
| Bank Customer Churn | 79.63% | 76.46% | 78.70% | 78.13% | 78.00% |
| Tech Students | 83.77% | 74.35% | 69.94% | 76.40% | 71.44% |

Table 4. Accuracy after applying GADIC

|  | SVM | KNN | LR | DT | NB |
|---|---|---|---|---|---|
| Cleveland Heart Disease | 81.11% | 90.00% | 86.66% | 78.88% | 87.77% |
| Indian Liver Patient | 84.70% | 91.76% | 82.94% | 77.64% | 77.05% |
| Pima Indian Diabetes | 85.28% | 94.37% | 84.84% | 83.11% | 85.71% |
| Employee Future Prediction | 69.69% | 87.10% | 76.00% | 86.74% | 76.00% |
| Telecom Churn Prediction | 81.99% | 91.23% | 85.40% | 79.90% | 76.68% |
| Bank Customer Churn | 89.86% | 91.96% | 89.16% | 78.90% | 88.70% |
| Tech Students | 87.70% | 87.44% | 73.34% | 76.61% | 72.46% |

From the results, all classifiers showed increases in accuracy by varying degrees across all datasets, with the exception of NB in only the Cleveland heart disease dataset. As can be seen in Table 5 and Table 6, which illustrate the difference in f1-score before and after applying GADIC, these findings were consistent for both accuracy and f1-score.

Table 5. F1-score before applying GADIC

|  | SVM | KNN | LR | DT | NB |
|---|---|---|---|---|---|
| Cleveland Heart Disease | 0.65 | 0.67 | 0.85 | 0.78 | 0.89 |
| Indian Liver Patient | 0.59 | 0.66 | 0.67 | 0.6 | 0.56 |
| Pima Indian Diabetes | 0.75 | 0.74 | 0.76 | 0.69 | 0.77 |
| Employee Future Prediction | 0.52 | 0.77 | 0.68 | 0.81 | 0.68 |
| Telecom Churn Prediction | 0.73 | 0.76 | 0.79 | 0.73 | 0.77 |
| Bank Customer Churn | 0.71 | 0.71 | 0.72 | 0.78 | 0.72 |
| Tech Students | 0.84 | 0.74 | 0.7 | 0.76 | 0.71 |

Table 6. F1-score after applying GADIC

|  | SVM | KNN | LR | DT | NB |
|---|---|---|---|---|---|
| Cleveland Heart Disease | 0.82 | 0.9 | 0.87 | 0.78 | 0.88 |
| Indian Liver Patient | 0.79 | 0.91 | 0.75 | 0.78 | 0.72 |
| Pima Indian Diabetes | 0.85 | 0.94 | 0.84 | 0.83 | 0.85 |
| Employee Future Prediction | 0.57 | 0.87 | 0.73 | 0.87 | 0.74 |
| Telecom Churn Prediction | 0.74 | 0.91 | 0.85 | 0.8 | 0.79 |
| Bank Customer Churn | 0.85 | 0.91 | 0.84 | 0.83 | 0.83 |
| Tech Students | 0.88 | 0.87 | 0.73 | 0.77 | 0.72 |

The extent to which GADIC enhanced the performance of the chosen classifiers in terms of accuracy is shown in Figure 3. According to the figure, KNN had the highest increase in performance when GADIC was applied, outperforming the other models in all datasets with remarkable improvement. SVM had the second highest increase in performance, while LR was the least affected classifier by GADIC by a small difference than NB and DT. LR's lowest increase is likely due to the nature and assumptions of the LR algorithm itself. LR classifier performance can be impacted by the presence of imbalanced data (Salas-Eljatib et al., 2018).

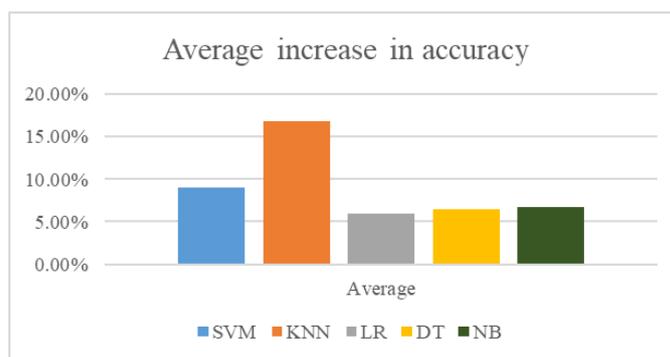

Figure 3. Average increase in accuracy per model

From the experiments, in most cases, DT and NB classifiers often stopped improving after a limited number of generations in GADIC, and the performance did not improve more with the increase in the number of generations. In contrast, LR took more generations to get its best results, but even then, the performance gain is less than that of other models. This indicates that this classifier is slightly influenced by GADIC compared with other classifiers.

As observed from the obtained results, the performance of the models was significantly influenced by the nature of the data. Greater improvements in performance after applying GADIC indicate that the types and nature of the data have a major influence on the results. On the contrary, lower improvements indicate that the data itself has little impact and that there may be other factors influencing the performance. Figure 4 shows the datasets with higher and lower average improvement in accuracy when applying GADIC.

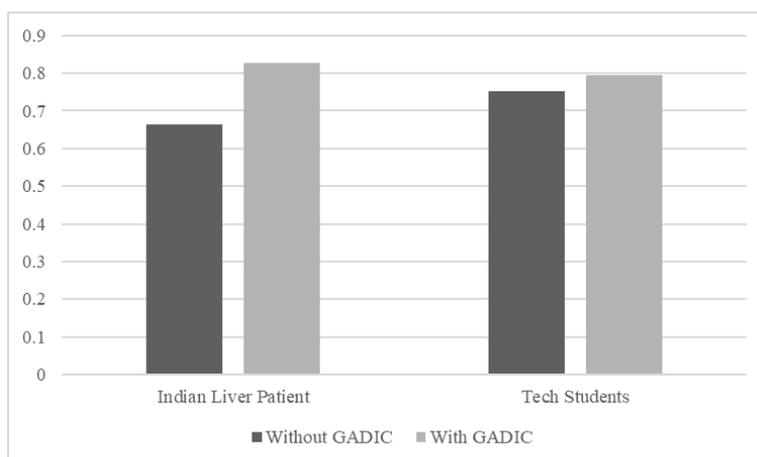

Figure 4. Datasets with higher and lower improvement in accuracy with GADIC

## 7. Conclusion

A novel classification algorithm, namely GADIC, that is based on GA and DI reformatting was proposed. GADIC was intended to be able to overcome some data-related issues that could prohibit the classifiers from improving their performance and hence increasing the performance of the classifiers. GADIC was tested on five ML classifiers using seven open-source datasets to assess its performance. The results demonstrated that GADIC improved the performance of the majority of the tested classifiers using various datasets in terms of size and type.

KNN had the most performance improvement when applying GADIC with an average increase of 16.79%, followed by SVM with 9.03% average increase. The other three classifiers (LR, DT, NB) had an average increase of 5.96%, 6.4% and 6.75% respectively, with LR showing the least improvement which is still considered to be an important achievement. The improvement in the models' performance with GADIC indicates that the nature of data is a factor influencing the performance of the classifiers.

Future studies should consider testing the GADIC approach on larger datasets, gauging its potential for enhancing performance of other ML and DL models, and refining the proposed algorithm by experimenting with and integrating different parameters.